\documentclass{article}

\PassOptionsToPackage{numbers, compress}{natbib}

\usepackage[final]{neurips_2020}

\usepackage[final]{neurips_2020}

\usepackage[utf8]{inputenc} %
\usepackage[T1]{fontenc}    %
\usepackage{hyperref}       %
\usepackage{url}            %
\usepackage{booktabs}       %
\usepackage{amsfonts}       %
\usepackage{nicefrac}       %
\usepackage{microtype}      %
\usepackage[dvipsnames]{xcolor}

\usepackage{times}
\usepackage{etoolbox}
\usepackage{amsmath,amssymb}
\usepackage{amsthm}
\usepackage{latexsym}
\usepackage{inconsolata}
\usepackage{bm}
\usepackage{empheq}
\usepackage{mathtools}
\usepackage{xspace}
\usepackage{enumitem}
\usepackage{setspace}
\usepackage{subcaption}
\usepackage{comment}
\usepackage{todonotes}
\usepackage{wrapfig}
\usepackage{sidecap}
\usepackage{algorithm}
\usepackage[noend]{algpseudocode}

\newcommand{\smap}{SparseMAP\xspace}
\newcommand*{\wrt}{\textit{w.\nobreak\hspace{.07em}r.\nobreak\hspace{.07em}t.}\@\xspace}
\newcommand*{\eg}{\textit{e.\nobreak\hspace{.07em}g.}\@\xspace}
\newcommand*{\ie}{\textit{i.\nobreak\hspace{.07em}e.}\@\xspace}

\newcommand{\eqnref}[1]{Eq.~\ref{#1}}
\newcommand{\secref}[1]{\S\ref{sec:#1}}
\newcommand{\figref}[1]{Fig.~\ref{#1}}

\let\log\relax
\let\exp\relax
\let\min\relax
\let\max\relax
\DeclareMathOperator*{\log}{\mathsf{log}}
\DeclareMathOperator*{\exp}{\mathsf{exp}}
\DeclareMathOperator*{\min}{\mathsf{min}}
\DeclareMathOperator*{\max}{\mathsf{max}}
\DeclareMathOperator{\KL}{\mathsf{KL}}

\DeclareMathOperator{\Cat}{\mathsf{Categorical}}

\DeclareMathOperator*{\softmax}{\mathsf{softmax}}
\DeclareMathOperator*{\smapop}{\mathsf{SparseMAP}}
\DeclareMathOperator*{\sparsemax}{\mathsf{sparsemax}}
\DeclareMathOperator*{\argmin}{\mathsf{argmin}}
\DeclareMathOperator*{\argmax}{\mathsf{argmax}}
\DeclareMathOperator*{\minimize}{\mathsf{minimize}}

\DeclareMathOperator*{\subjto}{\mathsf{s.t.}}
\DeclareMathOperator{\sparsemaxk}{\mathsf{sparsemax}_\text{$k$}}
\DeclareMathOperator{\topk}{\mathsf{top}_\text{$k$}}
\newcommand{\simplex}{\triangle}
\renewcommand\ss{s}
\newcommand\s{\bm{\ss}}

\newcommand{\reals}{\mathbb{R}}

\newcommand\ZZ{\mathcal{Z}}
\newcommand\LL{\mathcal{L}}
\renewcommand\AA{\bm{A}}
\newcommand\zsamp{z}
\newcommand\defeq\coloneqq
\newcommand\pv{\bm{\xi}}
\newcommand\pp{\xi}
\newcommand\g{\bm{g}}
\newcommand\dd{\operatorname{d}\!}

\newcommand{\EE}{\mathbb{E}}

\newcommand{\DP}[2]{\langle#1,#2\rangle}
\newcommand{\pfrac}[2]{\frac{\partial #1}{\partial #2}}
\newcommand\spacerule{\addlinespace[0.33em]}

\usepackage{soul}
\usepackage{marvosym}
\newcommand\markUnbabel{\Cancer}
\newcommand\markIT{\Leo}
\newcommand\markLUMLIS{\Neptune}
\newcommand\markILLC{\Virgo}
\newcommand\markIvI{\Libra}

\title{Efficient Marginalization of\\
Discrete and Structured Latent Variables via Sparsity}

\author{
Gon\c{c}alo M. Correia\textsuperscript{\markIT} \\
\href{mailto:goncalo.correia@lx.it.pt}{\tt goncalo.correia@lx.it.pt}
\And
Vlad Niculae\textsuperscript{\markIvI{}}%
\thanks{Work partially done while VN was at the Instituto de Telecomunica\c{c}\~oes, Lisbon.}%
\\
\href{mailto:vlad@vene.ro}{\tt vlad@vene.ro} \\
\And
Wilker Aziz\textsuperscript{\markILLC{}} \\
\href{mailto:w.aziz@uva.nl}{\tt w.aziz@uva.nl}\\
\And
Andr\'e F.~T. Martins\textsuperscript{\markIT{} \markLUMLIS{}\markUnbabel{}} \\
\href{mailto:andre.t.martins@tecnico.ulisboa.pt}{\tt andre.t.martins@tecnico.ulisboa.pt}\\
[1ex]
\textsuperscript{\markIT{}}Instituto de Telecomunica\c{c}\~oes, Lisbon, Portugal\\
\textsuperscript{\markLUMLIS{}}LUMLIS (Lisbon ELLIS Unit), Instituto Superior T\'ecnico, Lisbon, Portugal\\
\textsuperscript{\markUnbabel}Unbabel, Lisbon, Portugal\\
\textsuperscript{\markILLC}ILLC, University of Amsterdam, The Netherlands\\
\textsuperscript{\markIvI}IvI, University of Amsterdam, The Netherlands
}

\begin{document}
\maketitle
\begin{abstract}
Training neural network models with discrete (categorical or
structured) latent variables can be computationally challenging, due
to the need for marginalization over large or combinatorial sets. To
circumvent this issue, one typically resorts to sampling-based
approximations of the true marginal, requiring noisy gradient
estimators (e.g., score function estimator) or continuous relaxations
with lower-variance reparameterized gradients (e.g., Gumbel-Softmax).
In this paper, we propose a new training strategy which replaces
these estimators by an exact yet efficient marginalization. To
achieve this, we parameterize discrete distributions over latent
assignments using differentiable sparse mappings: sparsemax and its
structured counterparts. In effect, the support of these
distributions is greatly reduced, which enables efficient
marginalization. We report successful results in three tasks covering
a range of latent variable modeling applications: a semisupervised
deep generative model, a latent communication game, and a generative
model with a bit-vector latent representation. In all cases, we
obtain good performance while still achieving the practicality of
sampling-based approximations.
\end{abstract}

\section{Introduction}
\label{sec:intro}

Neural latent variable models are powerful and expressive tools for
finding patterns in high-dimensional data, such as images or text
\citep{Kim2018,Kingma+2014:VAE,RezendeEtAl14VAE}. Of particular
interest are \emph{discrete} latent variables, which can recover
categorical and structured encodings of hidden aspects of the data,
leading to compact representations and, in some cases, superior
explanatory power~\citep{titov2008joint, Bastings2019}. However, with
discrete variables, training can become challenging, due to the need
to compute a gradient of a large sum over all possible latent
variable assignments, with each term itself being potentially
expensive. This challenge is typically tackled by estimating the
gradient with Monte Carlo methods~\citep[MC;][]{mohamed2019monte},
which rely on sampling estimates. The two most common strategies for
MC gradient estimation are the score function
estimator~\citep[SFE;][]{rubinstein1976monte,paisley2012variational},
which suffers from high variance, or surrogate methods that rely on
the continuous relaxation of the latent variable, like
straight-through \citep{STE} or Gumbel-Softmax
\citep{Concrete,GumbelSoftmax} which potentially reduce variance but
introduce bias and modeling assumptions.

In this work, we take a step back and ask: Can we avoid sampling
entirely, and instead deterministically evaluate the sum with less
computation? To answer affirmatively, we propose an alternative
method to train these models by parameterizing the discrete
distribution with {\bf sparse
mappings}\,---\,sparsemax~\citep{martins2016softmax} and two
structured counterparts, \smap~\citep{niculae2018sparsemap} and a
novel mapping top-$k$ sparsemax. Sparsity implies that some
assignments of the latent variable are entirely ruled out. This leads
to the corresponding terms in the sum evaluating trivially to zero,
allowing us to disregard potentially expensive computations.

\paragraph{Contributions.} We introduce a general strategy for
learning deep models with discrete latent variables that hinges on
learning a sparse distribution over the possible assignments. In the
unstructured categorical case our strategy relies on the sparsemax
activation function, presented in~\secref{categorical}, while in the
structured case we propose two strategies, \smap and top-$k$
sparsemax, presented in~\secref{structured}. Unlike existing
approaches, our strategies involve neither MC estimation nor any
relaxation of the discrete latent variable to the continuous space.
We demonstrate our strategy on three different applications: a
semisupervised generative model, an emergent communication game, and
a bit-vector variational autoencoder. We provide a thorough analysis
and comparison to MC methods, and\,---\,when feasible\,---\,to exact
marginalization. Our approach is consistently a top performer,
combining the accuracy and robustness of exact marginalization with
the efficiency of single-sample estimators.\footnote{Code is publicly
available at
\href{https://github.com/deep-spin/sparse-marginalization-lvm}{\tt
https://github.com/deep-spin/sparse-marginalization-lvm}}

\paragraph{Notation.} We denote scalars, vectors, matrices, and sets
as $a$, $\bm{a}$, $\bm{A}$, and $\mathcal{A}$, respectively. The
indicator vector is denoted by $\bm{e}_{i}$, for which every entry is
zero, except the $i$\textsuperscript{th}, which is 1. The simplex is
denoted $\simplex^K \coloneqq \{ \pv \in \reals^K~:~\DP{\bm{1}}{\pv}
= 1, \pv \geq \bm{0}\}$. $\mathbb H(p)$ denotes the Shannon entropy
of a distribution $p(z)$, and $\KL\left[ p || q \right]$ denotes the
Kullback-Leibler divergence of $p(z)$ from $q(z)$. The number of
non-zeros of a sequence $z$ is denoted $\|z\|_0 \defeq |\{t: z_t \neq
0 \}|$. Letting $z \in \ZZ$ be a random variable, we write the
expectation of a function $f:\ZZ \rightarrow \mathbb{R}$ under
distribution $p(z)$ as $\mathbb{E}_{p(z)}[f(z)]$.

\section{Background}\label{sec:background}

We assume throughout a latent variable model with observed variables
$x \in \mathcal{X}$ and latent stochastic variables $z \in \ZZ$. The
overall fit to a dataset $\mathcal D$ is $\sum_{x \in \mathcal{D}}
\LL_x(\theta)$, where the loss of each observation,
\begin{equation}\label{eq:fit}
	\mathcal{L}_{x}(\theta) =
    \mathbb E_{\pi(z \mid x, \theta)}
    \left[ \ell(x, z; \theta)\right] =
    \sum_{z \in \mathcal Z} \pi(z | x, \theta)~\ell(x, z; \theta) ~,
\end{equation}
is the expected value of a downstream loss $\ell(x,z;\theta)$ under a
probability model $\pi(z|x,\theta)$ of the latent variable; in other
words, the latent variable $z$ is {\it marginalized} to compute this loss.
To model complex data, one parameterizes both the downstream loss and
the distribution over latent assignments using neural networks, due
to their flexibility and capacity~\cite{Kingma+2014:VAE}.

In this work, we study \textbf{discrete} latent variables, where
$|\ZZ|$ is finite, but possibly very large. One example is when
$\pi(z|x,\theta)$ is a categorical distribution, parameterized by a
vector $\pv \in \simplex^{|\ZZ|}$. To obtain $\pv$, a neural network
computes a vector of scores $\s \in \mathbb{R}^{|\ZZ|}$, one score
for each assignment, which is then mapped to the probability simplex,
typically via $\pv = \softmax(\s)$. Another example is when $\ZZ$ is
a structured (combinatorial) set, such as $\ZZ = \{0, 1\}^D$. In this
case, $|\ZZ|$ grows exponentially with $D$ and it is infeasible to
enumerate and score all possible assignments. For this structured
case, scoring assignments involves a decomposition into parts, which
we describe in \secref{structured}.

Training such models requires summing the contributions of all
assignments of the latent variable, which involves as many as $|\ZZ|$
evaluations of the downstream loss. When $\ZZ$ is not too large, the
expectation may be evaluated explicitly, and learning can proceed
with exact gradient updates. If $\ZZ$ is large, and/or if $\ell$ is
an expensive computation, evaluating the expectation becomes
prohibitive. In such cases, practitioners typically turn to MC
estimates of $\nabla_\theta \LL_x(\theta)$ derived from latent
assignments sampled from $\pi(z|x,\theta)$. Under an appropriate
learning rate schedule, this procedure converges to a local optimum
of $\mathcal L_x(\theta)$ as long as gradient estimates are unbiased
\citep{robbins1951stochastic}. Next, we describe the two current main
strategies for MC estimation of this gradient. Later, in
\S\ref{sec:categorical}--\ref{sec:structured}, we propose our
\textbf{deterministic} alternative, based on sparsifying $\pi(z| x,
\theta)$.

\paragraph{Monte Carlo gradient estimates.} Let $\theta=(\theta_\pi,
\theta_\ell)$, where $\theta_\pi$ is the subset of weights that $\pi$
depends on, and $\theta_\ell$ the subset of weights that $\ell$
depends on. Given a sample $\zsamp \sim \pi(z| x,\theta_\pi)$, an
unbiased estimator of the gradient for \eqnref{eq:fit} \wrt
$\theta_\ell$ is $\nabla_{\theta_\ell} \mathcal L_x(\theta) \approx
\nabla_{\theta_\ell} \ell(x, \zsamp; \theta_\ell)$. Unbiased
estimation of $\nabla_{\theta_\pi} \mathcal L_x(\theta)$ is more
difficult, since $\theta_\pi$ is involved in the sampling of $\zsamp$,
but can be done with SFE
\citep{rubinstein1976monte,paisley2012variational}:
$\nabla_{\theta_\pi} \mathcal L_x(\theta) \approx \ell(x, \zsamp;
\theta_\ell) ~ \nabla_{\theta_\pi} \log \pi(\zsamp | x, \theta_\pi)$,
also known as \textsc{reinforce} \citep{Williams1992}. The SFE is
powerful and general, making no assumptions on the form of $z$ or
$\ell$, requiring only a sampling oracle and a way to assess
gradients of $\log \pi(\zsamp | x, \theta_\pi)$. However, it comes
with the cost of high variance. Making the estimator practically
useful requires variance reduction techniques such as
baselines~\citep{Williams1992,MuProp} and control
variates~\citep{CV2013,REBAR,RELAX}. Variance reduction can also be
achieved with Rao-Blackwellization techniques such as sum and
sample~\citep{casella1996rao,BBVI14,RB19}, which marginalizes an
expectation over the top-$k$ elements of $\pi(z| x,\theta_\pi)$ and
takes a sample estimate from the complement set.

\paragraph{Reparameterization trick.} For continuous latent variables,
low-variance pathwise gradient estimators can be obtained by
separating the source of stochasticity from the sampling parameters,
using the so-called \emph{reparameterization trick}
\citep{Kingma+2014:VAE,RezendeEtAl14VAE}. For discrete latent
variables, reparameterizations can only be obtained by introducing a
step function like $\argmax$, which has null gradients almost everywhere.
Replacing the gradient of $\argmax$ with a nonzero surrogate like the identity
function, known as Straight-Through \citep{STE}, or with the gradient of $\softmax$, known
as \emph{Gumbel-Softmax} \citep{Concrete,GumbelSoftmax}, leads to a
biased estimator that can still perform well in practice. Continuous
relaxations like Straight-Through and Gumbel-Softmax are only
possible under a further modeling assumption that $\ell$ is defined
continuously (thus differentiably) in a neighbourhood of the
indicator vector $\bm{z} = \bm{e}_{\zsamp}$ for every $z \in \ZZ$. In
contrast, both SFE-based methods as well as our approach make no such
assumption.

\section{Efficient Marginalization via Sparsity}
\label{sec:categorical}

The challenge of computing the exact expectation in \eqnref{eq:fit}
is linked to the need to compute a sum with a large number of terms.
This holds when the probability distribution over latent assignments
is {\it dense} ({\it i.e.}, every assignment $z \in \ZZ$ has non-zero
probability), which is indeed the case for most parameterizations of
discrete distributions. Our proposed methods hinge on {\it
sparsifying} this sum.

Take the example where $\mathcal Z = \{1, \ldots, K\}$, with a neural
network predicting from $x$ a $K$-dimensional vector of real-valued
scores $\s = \g(x; \theta)$, such that $s_z$ is the score of
$z$.\footnote{Not to be confused with ``score function,'' as in SFE,
which refers to the gradient of the log-likelihood.} The traditional
way to obtain the vector $\pv$ parameterizing $\pi(z|x,\theta)$ is
with the softmax transform, \ie $\pv = \softmax(\s)$. Since this
gives $\pi(z|x,\theta) \propto \exp(s_z)$, the expectation in
\eqnref{eq:fit} depends on $\ell(x, z; \theta)$ for every possible
$z$.

We rethink this standard parameterization, proposing a \textbf{sparse}
mapping from scores to the simplex. In particular, we substitute
softmax by sparsemax~\citep{martins2016softmax}:
\begin{equation}\label{eq:sparsemax}
    \sparsemax(\s) \coloneqq
    \argmin_{\pv \in \simplex^K} \| \pv - \s \|_2^2 ~.
\end{equation}
Like softmax, sparsemax is differentiable and has efficient forward and backward
passes~\citep{Held1974,martins2016softmax}, described in detail in
App.~\ref{app:sparsemax}; the backward pass is essential in our use case. Since
\eqnref{eq:sparsemax} is the Euclidean projection operator onto the probability
simplex, and solutions can hit the boundary, sparsemax is likely to assign {\bf
probabilities of exactly zero}; in contrast, softmax is always dense.

Our main insight is that with a sparse parameterization of $\pi$, we
can compute the expectation in \eqnref{eq:fit} evaluating $\ell(x, z;
\theta)$ only for assignments $z \in \bar\ZZ \defeq \{z : \pi(z | x,
\theta) > 0\}$. This leads to a powerful alternative to MC
estimation, which requires fewer than $|\ZZ|$ evaluations of $\ell$,
and which strategically\,---\,yet deterministically\,---\,selects
which assignments $\bar\ZZ$ to evaluate $\ell$ on. Empirically, our
analysis in \secref{applications} reveals an adaptive behavior of
this sparsity-inducing mechanism, performing more loss evaluations in
early iterations while the model is uncertain, and quickly reducing
the number of evaluations, especially for unambiguous data points.
This is a notable property of our learning strategy: In contrast, MC
estimation cannot decide when an ambiguous data point may require
more sampling for accurate estimation; and directly evaluating
\eqnref{eq:fit} with the dense $\pv$ resulting from a softmax
parameterization never reduces the number of evaluations required,
even for simple instances.

\section{\label{sec:structured}Structured Latent Variables}

While the approach described in \secref{categorical} theoretically
applies to any discrete distribution, many models of interest involve
structured (or combinatorial) latent variables. In this section, we
assume $z$ can be represented as a {\it bit-vector}---\ie a random
vector of discrete binary variables $\bm{a}_{z} \in \{0, 1\}^D$. This
assignment of binary variables may involve global factors and
constraints (\eg tree constraints, or budget constraints on the
number of active variables, \ie $\sum_i [\bm{a}_{z}]_i \le B$, where
$B$ is the maximum number of variables allowed to activate at the
same time). In such structured problems, $|\ZZ|$ increases
exponentially with $D$, making exact evaluation of $\ell(x, z;
\theta)$ prohibitive, even with sparsemax.

Structured prediction typically handles this combinatorial explosion
by parameterizing scores for individual binary variables and
interactions within the global structured configuration, yielding a
compact vector of \textbf{variable scores} $\bm{t} = \g(x; \theta)
\in \mathbb{R}^D$ (\eg, log-potentials for binary attributes), with
$D \ll |\ZZ|$. Then, the score of some global configuration $z \in
\ZZ$ is $s_z \defeq \DP{\bm{a}_z}{\bm{t}}$. The variable scores
induce a unique Gibbs distribution over structures, given by
$\pi(z|x,\theta) \propto \exp(\DP{\bm{a}_{z}}{\bm{t}})$.
Equivalently, defining $\bm{A} \in \mathbb{R}^{D \times |\ZZ|}$ as
the matrix with columns $\bm{a}_{z}$ for all $z \in \ZZ$, we consider
the discrete distribution parameterized by $\softmax(\s)$, where
$\s=\bm{A}^\top \bm{t}$. (In the unstructured case, $\bm{A}=\bm{I}$.)

In practice, however, we cannot materialize the matrix $\AA$ or the
global score vector $\bm{s}$, let alone compute the softmax and the
sum in \eqnref{eq:fit}. The SFE, however, can still be used, provided
that exact sampling of $z\sim\pi(z | x, \theta)$ is feasible, and
efficient algorithms exist for computing the normalizing constant
$\sum_{z'}\exp(\DP{\bm{a}_{z'}}{\bm{t}})$~\citep{WJ2008}, needed to
compute the probability of a given sample.

While it may be tempting to consider using sparsemax to avoid the
expensive sum in the exact expectation, this is prohibitive too: solving the
problem in \eqnref{eq:sparsemax} still requires explicit manipulation
of the large vector $\bm{s} \in \mathbb{R}^{|\ZZ|}$, and even if we
could avoid this, in the worst case ($\bm{s}=\bm{0}$) the resulting
sparsemax distribution would still have exponentially large support.
Fortunately, we show next that it is still possible to develop
sparsification strategies to handle the combinatorial explosion of
$\ZZ$ in the structured case. We propose two different methods to
obtain a sparse distribution $\pv$ supported only over a bounded-size
subset of $\ZZ$: top-$k$ sparsemax (\S\ref{sec:topksparse}) and \smap
(\S\ref{sec:smap}).

\subsection{\label{sec:topksparse}Top-{\boldmath $k$} Sparsemax}

Recall that the sparsemax operator (\eqnref{eq:sparsemax}) is simply
the Euclidean projection onto the $|\ZZ|$-dimensional probability
simplex. While this projection has a propensity to be sparse, there
is no upper bound on the number of non-zeros of the resulting
distribution. When $\ZZ$ is large, one possibility is to add a
cardinality constraint $\|\pv\|_0 \le k$ for some prescribed $k \in
\mathbb{N}$. The resulting problem becomes
\begin{equation}\label{eq:topk_sparsemax}
    \sparsemaxk(\s) \coloneqq
    \argmin_{\pv \in \simplex^{|\ZZ|}, \, \|\pv\|_0 \le k} \| \pv - \s \|_2^2,
\end{equation}
which is known as a \emph{sparse projection onto the simplex} and has
been studied in detail by \citet{kyrillidis2013sparse} and used to
smooth structured prediction
losses~\citep{NIPS2018_7726,blondel2020}. Remarkably, while this is a
non-convex problem, its solution $\pv^\star$ can be written as a
composition of two functions: a top-$k$ operator $\topk:
\mathbb{R}^{|\ZZ|} \rightarrow \mathbb{R}^{|\ZZ|}$, which returns a
vector identical to its input but where all the entries not among the
$k$ largest ones are masked out (set to $-\infty$), and the
$k$-dimensional sparsemax operator.

Formally, $\sparsemaxk = \sparsemax(\topk(\s))$. Being a composition
of operators, its Jacobian becomes a product of matrices and hence
simple to compute (the Jacobian of $\topk$ is a diagonal matrix whose
diagonal is a multi-hot vector indicating the top-$k$ elements of
$\s$).

To apply the top-$k$ sparsemax to a large or combinatorial set $\ZZ$,
all we need is a primitive to compute the top-$k$ entries of
$s$---this is available for many structured problems (for example,
sequential models via $k$-best dynamic programming) and, when $\ZZ$
is the set of joint assignments of $D$ discrete binary variables, it
can be done with a cost $\mathcal{O}(kD)$.

After enumerating this set, we parameterize $\pi(z|x,\theta)$ by
applying sparsemax to that top-$k$, with a
computational cost $\mathcal{O}(k)$. Note that {\bf this method is
identical to sparsemax whenever $\|\sparsemax(\s)\|_0 \le k$}: if
during training the model learns to assign a sparse distribution to
the latent variable, we are effectively using a sparsemax
parameterization as presented in \secref{categorical} with cheap
computation. In fact, the solution of \eqnref{eq:topk_sparsemax}
gives us a certificate of optimality whenever $\|\pv^\star\|_0 < k$.

\subsection{\label{sec:smap}\smap}
A second possibility to obtain efficient summation over a
combinatorial space without imposing any constraints on $\ell(x, z;
\theta)$ is to use \smap \citep{niculae2018sparsemap, sparsemapcg}, a
structured extension of sparsemax:
\begin{equation}\label{eq:sparsemap}
\smapop(\bm{t}) \defeq \argmin_{\pv \in \simplex^{|\ZZ|}}
\|\bm{A}\pv - \bm{t}\|_2^2,
\end{equation}
\smap has been used successfully in discriminative latent models to
model structures such as trees and matchings, and
\citet{niculae2018sparsemap} apply an active set algorithm for
evaluating it and computing gradients efficiently, requiring only a
primitive for computing $\argmax_{z \in \ZZ} \langle \bm{a}_z,
\bm{t}\rangle$.
(We detail the algorithm in App.~\ref{app:activeset}).
While the $\argmin$ in \eqref{eq:sparsemap} is
generally not unique,
Carath\'eodory's theorem guarantees that
solutions with support size at most $D+1$ exist,
and the active set algorithm
enjoys
linear and finite convergence to a very sparse optimal distribution.
Crucially, \eqref{eq:sparsemap} has a solution $\bm{\xi}^\star$ such
that the set $\bar\ZZ = \left\{z \in \ZZ \mid \xi^\star_z >
0\right\}$ grows only linearly with $D$, and therefore $|\bar\ZZ| \ll
|\ZZ|$. Therefore, assessing the expectation in \eqnref{eq:fit} only
requires evaluating $|\bar\ZZ| = \mathcal{O}(D)$ terms.

\section{\label{sec:applications}Experimental Analysis}

We next demonstrate the applicability of our proposed strategies by
tackling three tasks: a deep generative model with semisupervision
(\secref{gen}), an emergent communication two-player game over a
discrete channel (\secref{comm}), and a variational autoencoder with
latent binary factors (\secref{bernvae}). We describe any further
architecture and hyperparameter details in App.~\ref{app:details}.

\subsection{Semisupervised Variational Auto-encoder (VAE)}\label{sec:gen}

\begin{figure}[t]
\begin{minipage}[c]{0.52\linewidth}
\vspace{0pt}
\centering
\includegraphics[width=.9\textwidth]{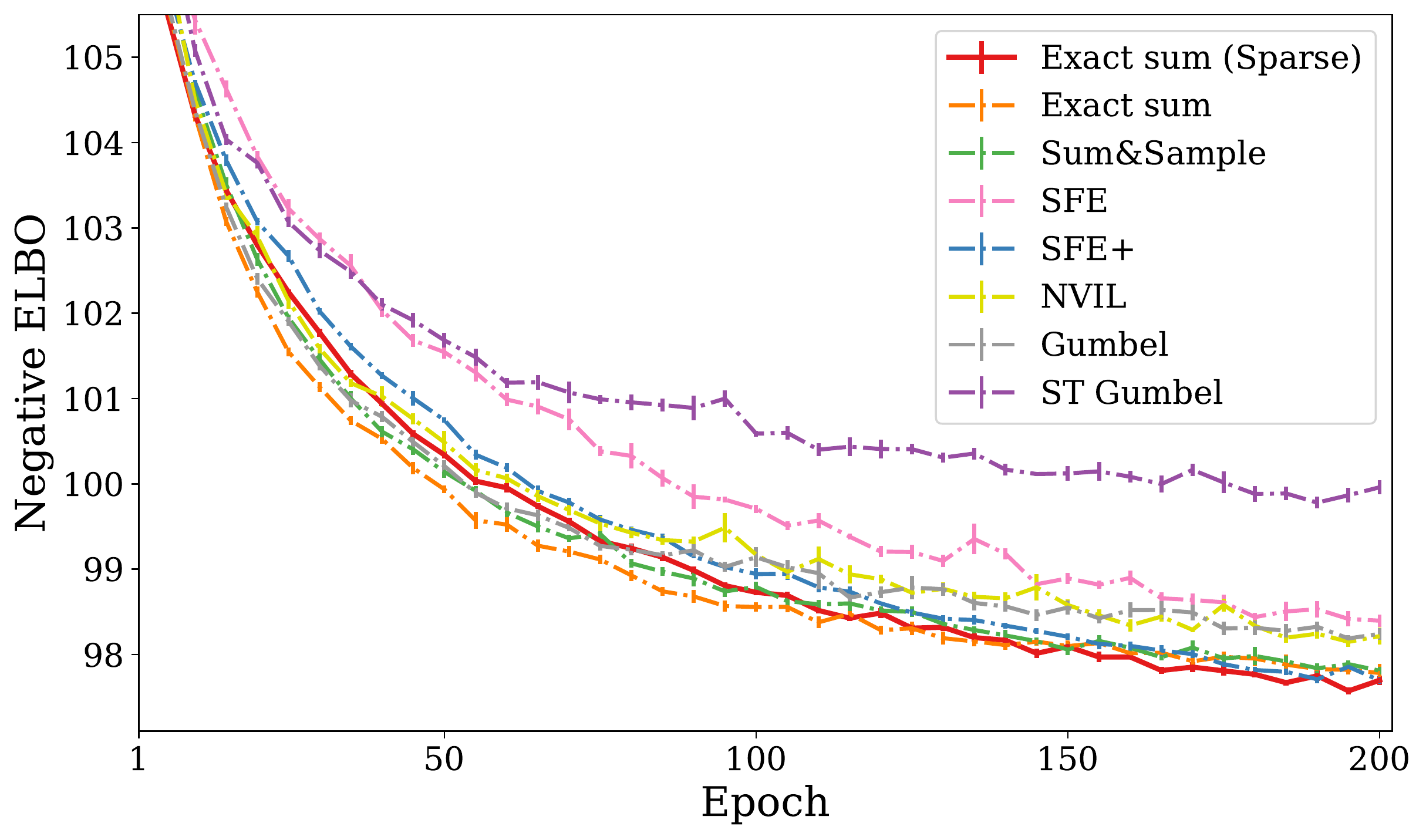}
\vspace{0pt}
\end{minipage}
\begin{minipage}[c]{0.45\linewidth}
\vspace{0pt}
\centering
\small
    \begin{tabular}{lrr}
    \toprule
    Method &
    Accuracy (\%)
    & Dec. calls\\
    \midrule

\multicolumn{3}{l}{\emph{Monte Carlo}} \\
    SFE
    & $94.75${\tiny\color{gray}$\pm .002$} & $1$ \\
    SFE$+$
    & $96.53${\tiny\color{gray}$\pm .001$}  & $2$  \\
    NVIL
    & $96.01${\tiny\color{gray}$\pm .002$}  & $1$  \\
    Sum\&Sample
    & $96.73${\tiny\color{gray}$\pm .001$}  & $2$  \\
    Gumbel
    & $95.46${\tiny\color{gray}$\pm .001$}  & $1$  \\
    ST Gumbel
    & $86.35${\tiny\color{gray}$\pm .006$}  & $1$  \\
\spacerule
\multicolumn{3}{l}{\emph{Marginalization}} \\
    Dense
    & $96.93${\tiny\color{gray}$\pm .001$}  & $10$  \\
    Sparse {\small \color{gray}{(proposed)}}
    & $96.87${\tiny\color{gray}$\pm .001$}  & $1.01${\tiny\color{gray}$\pm 0.01$}  \\
    \bottomrule
    \end{tabular}
\vspace{5pt}
\end{minipage}
\caption{\label{fig:ssvaeelbo}Semisupervised VAE on MNIST.
Left: Learning curves (test).
Right: Average test results and standard errors over 10 runs.}
\end{figure}

We consider the semisupervised VAE of \citet{KingmaEtAl2014SSVAE},
which models the joint probability $p(z,h,x|\phi)=p(z)p(h)p(x|z,h)$, where
$x$ is an observation (an MNIST image), $h$ is a continuous latent
variable with a $n$-dimensional standard Gaussian prior, and $z$ is a
discrete random variable with a uniform prior over $K$ categories.
The marginal $p(x | \phi) = \sum_{z=1}^K \int_h p(x | z, h,
\phi)p(h)p(z) \dd h$ is intractable, due to the marginalization of $h \in
\mathbb R^n$. For a fixed $h$ (\eg, sampled), marginalizing $z$
requires $K$ calls to the decoder, which can be costly depending on
the decoder's architecture.

To circumvent the need for the marginal likelihood,
\citet{KingmaEtAl2014SSVAE} use variational inference
\citep{Jordan+1999:VI} with an approximate posterior $\pi(z|x,
\theta_\pi)q(h|z,x, \lambda)$. This trains a
classifier $\pi(z|x, \theta_\pi)$ along with the generative model. In
\citep{KingmaEtAl2014SSVAE}, $h$ is sampled with a
reparameterization, and the expectation over $z$ is computed in
closed-form, that is, assessing all $K$ terms of the sum for a
sampled $h$. Under the notation in \secref{background}, we let $\theta_\ell = \{\lambda, \phi\}$ and define
\begin{equation}
    \ell(x, z; \theta_\ell) \defeq
    - \mathbb E_{q(h|z,  \lambda)}\left[ \log p(x \mid z, h, \phi) \right] -
    \log \frac{p(z)}{\pi(z \mid x, \theta_\pi)} +
    \KL\left[q(h \mid x, z, \lambda) \,\,\|\,\, p(h)\right],
    \label{eq:elbonotation}
\end{equation}
which turns \eqnref{eq:fit} into the (negative) evidence lower bound
(ELBO). To update $q(h | x, z, \lambda)$, we use the
reparameterization trick to obtain gradients through a sampled $h$.
For $\pi(z | x, \theta_\pi)$, we may still explicitly marginalize over
each possible assignment of $z$, but this has a multiplicative cost
on $K$. As an alternative, we parameterize $\pi(z|x,
\theta_\pi)$ with a sparse mapping, comparing it to the original
formulation and with stochastic gradients based on SFE and continuous
relaxations of $z$.

\paragraph{Data and architecture.} We evaluate this model on the
MNIST dataset \citep{lecun1998gradient}, using 10\% of labeled data,
treating the remaining data as unlabeled. For the parameterization of
the model components we follow the architecture and training
procedure used in \citep{RB19}. Each model was trained for 200
epochs.

\paragraph{Comparisons.} Our proposal's key ingredient is sparsity,
which permits exact marginalization and a deterministic gradient. To
investigate the impact of sparsity alone, we report a comparison
against the exact marginalization over the entire support $\ZZ$ using
a dense softmax parameterization. To investigate the impact of
deterministic gradients, we compare to stochastic gradients. Unbiased
gradient estimators: \emph{(i)} SFE with a moving average baseline;
\emph{(ii)} SFE with a self-critic
baseline~\citep[SFE+;][]{rennie2017self}, that is, we use $\log
p(x|z', h, \phi)$ as baseline, where $z' \sim \pi(z|x, \theta_\pi)$
is an independent sample; \emph{(iii)} NVIL \citep{mnih2014neural}
with a learned baseline (we train a MLP to predict the learning
signal by minimizing mean squared error);
and \emph{(iv)} sum-and-sample, a Rao-Blackwellized
version of SFE~\citep{RB19}. Biased gradient estimators: \emph{(v)}
Gumbel-Softmax, which relaxes the random variable to the simplex, and
\emph{(vi)} ST Gumbel-Softmax, which discretizes the relaxation in
the forward pass, but ignores the discretization function in the
backward pass.\footnote{\label{note:GS} For Gumbel-Softmax (with and
without ST), we follow \citet{GumbelSoftmax} and substitute
$\KL(\pi(z|x, \theta_\pi) \| p(z))$ in the ELBO by the $\KL$
divergence of $\Cat(\softmax(\s))$ from a discrete uniform prior.
Strictly speaking this means the objective is not a proper ELBO and
its relationship to an ELBO is unclear \citep[Appendix
C.2]{Concrete}.}

\paragraph{Results and discussion.}

In \figref{fig:ssvaeelbo}, we see that our proposed sparse
marginalization approach performs just as well as its dense
counterpart, both in terms of ELBO and accuracy. However, by
inspecting the number of times each method calls the decoder for
assessments of $p(x|z, h,\phi)$, we can see that the effective
support of our method is much smaller\,---\,sparsemax-parameterized
posteriors get very confident, and mostly require one, and sometimes
two, calls to the decoder. Regarding the Monte Carlo methods, the
continuous relaxation done by Gumbel-Softmax underperformed all the
other methods, with the exception of SFE with a moving average. While
SFE+ and Sum\&Sample are very strong performers, they will always
require throughout training the same number of calls to the decoder
(in this case, two). On the other hand, sparsemax makes a small
number of decoder calls not due to a choice in hyperparameters but
thanks to the model converging to only using a small support, which
can endow this method with a lower number of computations as it
becomes more confident.

\subsection{Emergent Communication Game}\label{sec:comm}

Emergent communication studies how two agents can develop a
communication protocol to solve a task
collaboratively~\citep{kirby2002natural}. Recent work used neural
latent variable models to train these agents via a ``collaborative
game'' between
them~\citep{lewis1969convention,Lazaridou2017,Havrylov2017,
jorge2016learning, foerster2016learning, sukhbaatar2016learning}. In
\citep{Lazaridou2017}, one of the agents (the \emph{sender}) sees an
image $v_y$ and sends a single symbol message $z$ chosen from a set
$\mathcal{Z}$ (the \emph{vocabulary}) to the other agent (the
\emph{receiver}), who needs to choose the correct image $v_y$ out of
a collection of images $\mathcal{V} = \left\{ v_1, \dots, v_C
\right\}$.\footnote{\citet{Lazaridou2017} lets the sender see the
full set $\mathcal{V}$. In contrast, we follow \citep{Havrylov2017}
in showing only the correct image $v_y$ to the sender. This makes the
game harder, as the message $z$ needs to encode a good
``description'' of the correct image $v_y$ instead of encoding only
its differences from $\mathcal{V}\setminus \{v_y\}$.} They found that
the messages communicated this way can be correlated with broad
object properties amenable to interpretation by humans. In our
framework, we let $x = (\mathcal{V}, y)$ and define $\ell (x, z;
\theta) \coloneqq -\log p(y \mid \mathcal{V}, z, \theta_\ell)$ and
$\pi (z \mid x, \theta) \coloneqq p(z \mid v_y, \theta_\pi)$, where
$p(y \mid \mathcal{V}, z, \theta_\ell)$ corresponds to the sender and
$p(z \mid v_y, \theta_\pi)$ to the receiver. Following
\citet{Lazaridou2017}, we add an entropy regularization of $\pi (z
\mid x, \theta)$ to the loss, with a coefficient as an
hyperparameter~\citep{Mnih2016}.

\paragraph{Data and architecture.} We follow the architecture
described in \citep{Lazaridou2017}. However, to make the game harder,
we increase the collection of images $|\mathcal{V}|$ as suggested by
\citep{Havrylov2017}; in our experiments, we increase it from 2 to
16. All methods are trained for 500 epochs.

\paragraph{Comparisons.} We compare our method to stochastic gradient
estimators as well as exact marginalization under a dense softmax
parameterization of $p (z \mid v_y, \theta_\pi)$. Again, we have
unbiased (SFE with moving average baseline, SFE+, and NVIL) and
biased (Gumbel-Softmax and ST Gumbel-Softmax) estimators. For SFE we
also experiment with a 0/1 loss, rather than negative log-likelihood
(NLL).

\begin{wraptable}[17]{R}{80mm}
    \caption{Emergent communication success test results,
    averaged across 10 runs. Random guess baseline $6.25\%$.}
    \label{tab:symbol}
        \begin{center}
        \begin{small}
        \begin{tabular}{lr@{~}lr}
        \toprule
        Method & \multicolumn{2}{c}{Comm. succ. (\%)}  & Dec. calls  \\
        \midrule
    {\emph{Monte Carlo}} & & & \\
        SFE (NLL)  & $33.05$&{\tiny\color{gray}$\pm 2.84$}  & $1$  \\
        SFE (0/1)  & $55.36$&{\tiny\color{gray}$\pm 2.92$}  & $1$  \\
        SFE$+$ (0/1)  & $44.32$&{\tiny\color{gray}$\pm 2.72$}  & $2$  \\
        NVIL  & $37.04$&{\tiny\color{gray}$\pm 1.61$}  & $1$  \\
        Gumbel     & $23.51$&{\tiny\color{gray}$\pm 16.19$}  & $1$  \\
        ST Gumbel  & $27.42$&{\tiny\color{gray}$\pm 13.36$}  & $1$  \\
    \spacerule
    \emph{Marginalization} & & & \\
        Dense & $93.37$&{\tiny\color{gray}$\pm 0.42$}  & $256$  \\
        Sparse {\small \color{gray}(proposed)}  &
            $93.35$&{\tiny\color{gray}$\pm 0.50$} &
            $3.13${\tiny\color{gray}$\pm 0.48$} \\
        \bottomrule
        \end{tabular}
        \end{small}
        \end{center}
\end{wraptable}

\paragraph{Results and discussion.}

Table~\ref{tab:symbol} shows the communication success (accuracy of
the receiver at picking the correct image $v_y$). While the
communication success for $|\mathcal{V}|=2$ in \citep{Lazaridou2017}
was close to perfect, we see that increasing $|\mathcal{V}|$ to 16
makes this game much harder to sampling-based approaches. In fact,
only the models that do explicit marginalization achieve close to
perfect communication in the test set. However, as $\ZZ$ increases,
marginalizing with a softmax parameterization gets computationally
more expensive, as it requires $|\ZZ|$ forward and backward passes on
the receiver. Unlike softmax, the model trained with sparsemax
gives a very small support, requiring on average only 3 decoder
calls. In fact, sparsemax
starts off dense while exploring, but quickly becomes very sparse
(App.~\ref{app:dec_calls}).

\subsection{Bit-Vector Variational Autoencoder}\label{sec:bernvae}

As described in \S\ref{sec:structured}, in many interesting problems,
combinatorial interactions and constraints make $\ZZ$ exponentially
large. In this section, we study the illustrative case of encoding
(compressing) images into a binary codeword $z$, by training a latent
bit-vector variational autoencoder~\citep{GumbelSoftmax,
mnih2014neural}. One approach for parameterizing the approximate
posterior is to use a Gibbs distribution, decomposable as a product
of independent Bernoulli, $q(z \mid x, \lambda) \propto
\exp(\DP{\bm{a}_z}{\bm{t}}) = \prod_{i=1}^D q(z_i \mid x, \lambda)$,
with each $z_i$ being a Bernoulli with parameter $t_i$, and $D$ being
the number of binary latent variables. While marginalizing over all the
possible $z \in \ZZ$ is intractable, drawing samples can be done
efficiently by sampling each component independently, and the entropy
has a closed-form expression.
This efficient sampling and entropy computation relies on an independence
assumption; in general, we may not have access to such efficient
computation.

Training this VAE to minimize the negative ELBO corresponds to
$\ell(x, z; \theta_\ell) \defeq - \log \frac{p(x, z | \phi)}{q(z | x,
\lambda)} $; we use a uniform prior $p(z)=1/|\ZZ|=1/{2^D}$. This
objective does not constrain $\pi(z| x,\theta_\pi) \defeq q(z \mid x,
\lambda)$ to the Gibbs parameterization, and thus to apply our
methods we will differ from it.

\paragraph{Top-{\boldmath $k$} sparsemax parameterization.} As pointed
out in \secref{structured}, we cannot explicitly handle the
structured sparsemax distribution $\pv = \sparsemax(\s)$, as it
involves a vector of dimension $2^D$. However, given $\bm{t}$, we can
efficiently find the $k$ largest configurations in time
$\mathcal{O}(kD)$, with the procedure described in
\S\ref{sec:topksparse}, and thus we can evaluate $\sparsemaxk(\s)$
efficiently.

\paragraph{\smap parameterization.} Another sparse alternative to the
intractable structured sparsemax, as discussed in
\secref{structured}, is \smap. In this case, we compute an optimal
distribution $\pv$ using the active set algorithm of
\citep{niculae2018sparsemap}, by using a maximization oracle which
can be computed in $\mathcal{O}(D)$:
\begin{equation}
\argmax_z \DP{\bm{a}_z}{\bm{t}} = z^\star \quad \text{s.t.}\quad
[\bm{a}_{z^\star}]_i = \begin{cases}
1,& t_i \geq 0 \\
0,& t_i < 0 \\
\end{cases}.
\end{equation}
Since SparseMAP can handle structured problems, we also experimented
with adding a \emph{budget constraint} to SparseMAP: this is done by
adding a constraint $\|\bm{z}\|_1 \le B$, where $B \le D$; we used
$b=\frac{D}{2}$. The budget constraint ensures the images are
represented with sparse codes, and the maximization oracle can be
computed in $\mathcal{O}(D \log D)$ as described in
App.~\ref{app:budget}.

We stress that, with both top-$k$ sparsemax and \smap parameterizations,
$z$ does not decompose into a product of independent
binary variables: this property is specific to the Gibbs parameterization.
However, since these new approaches induce a very sparse approximate posterior
$q$, we may compute the terms $\EE_{q(z\mid x, \lambda)} [\log p(x \mid z,
\phi)]$ and $\EE_{q(z\mid x, \lambda)} [\log q(z \mid x, \lambda)]$
explicitly.

\begin{figure}
    \centering
    \begin{minipage}[c]{0.52\linewidth}
    \vspace{0pt}
    \includegraphics[width=.95\textwidth]{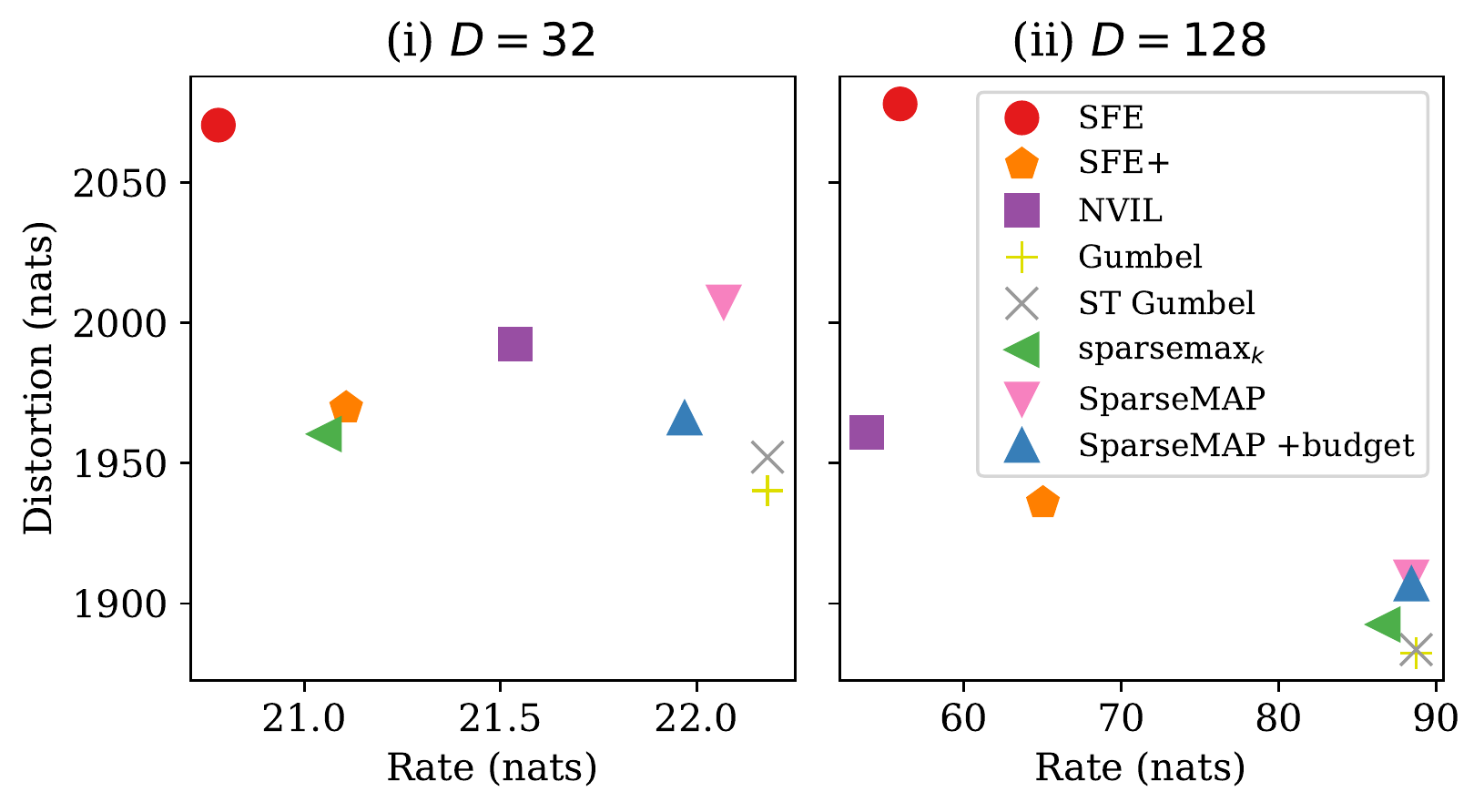}
    \vspace{0pt}
    \end{minipage}%
    \begin{minipage}[c]{0.45\linewidth}
    \vspace{0pt}
    \small
    \begin{tabular}{lrr}
        \toprule
        Method & $D=32$ & $D=128$\\
        \midrule
    \multicolumn{3}{l}{\emph{Monte Carlo}} \\
        SFE & $3.74$ & $3.77$  \\
        SFE$+$ & $3.61$ & $3.59$  \\
        NVIL & $3.65$ & $3.60$ \\
        Gumbel & $3.57$ & $3.49$  \\
        ST Gumbel & $3.53$ & $3.55$  \\
    \spacerule
    \multicolumn{3}{l}{\emph{Marginalization}} \\
        Top-$k$ sparsemax & $3.62$ & $3.61$  \\
        \smap & $3.72$ & $3.67$  \\
        \smap (w/ budget) & $3.64$ & $3.66$  \\
        \bottomrule
    \end{tabular}
    \vspace{5pt}
    \end{minipage}
    \caption{\label{fig:distortion}
        Test results for Fashion-MNIST.
        Left and middle: RD plots
        (the closer to the lower right corner, the better).
        Right: NLL in bits/dim (lower, the better).}
    \end{figure}

\paragraph{Data and architecture.} We use
Fashion-MNIST~\citep{fmnist}, consisting of 256-level grayscale
images $x \in {\{0, 1, \dots, 255\}}^{28\times 28}$. The decoder uses
an independent categorical distribution for each pixel, $p(x \mid z,
\phi) = \prod_{i=1}^{28} \prod_{j=1}^{28} p(x_{ij} \mid z, \phi)$.
For top-$k$ sparsemax, we choose $k=10$.

\begin{wrapfigure}[24]{r}{0.33\textwidth}
    \centering
    \includegraphics[width=0.33\textwidth]{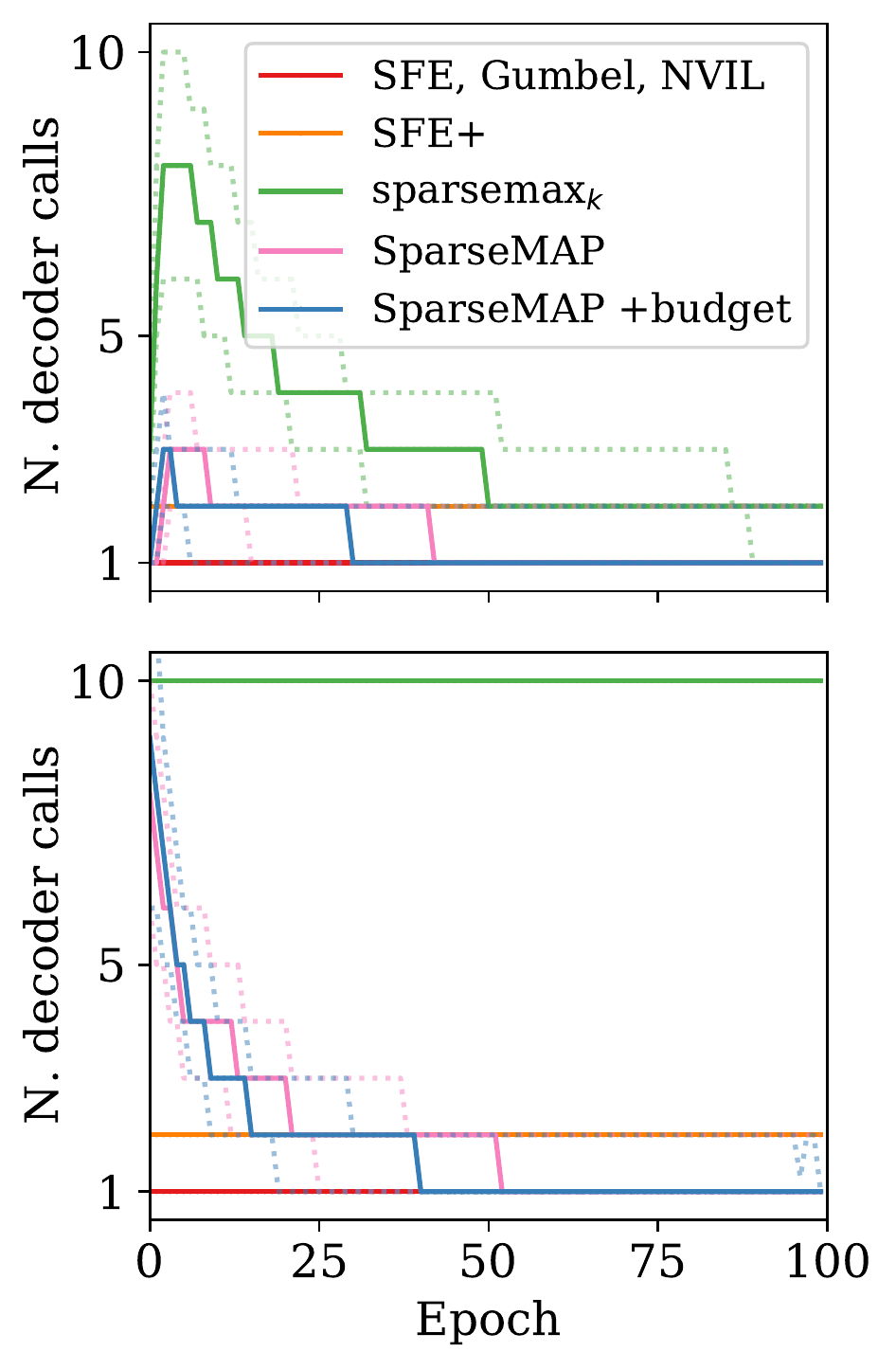}
    \caption{Bit vector VAE median and quartile decoder calls per
    epoch,
    $D=32$ (top) / $D=128$ (bottom).\label{fig:structcalls}}
\end{wrapfigure}

\paragraph{Comparisons.} This time, exact marginalization under a
dense parameterization of $q (z | x, \lambda)$ is truly intractable,
so we can only compare our method to stochastic gradient estimators.
We have unbiased SFE-based estimators (SFE with moving average
baseline, SFE+, and NVIL), and biased reparameterized gradient
estimators (Gumbel-Softmax and ST Gumbel-Softmax). As there is no
supervision for the latent code, we cannot compare the methods in
terms of accuracy or task success. Instead, we display the trained
models in the rate-distortion (RD) plane~\citep{Alemi2018}\footnote{Distortion is the
expected value of the reconstruction negative log-likelihood, while
rate is the average KL divergence from the prior to the approximate
posterior.} and also report bits-per-dimension of $x$, estimated with
importance sampling (App.~\ref{app:is}), on held-out data.

\paragraph{Results and discussion.} \figref{fig:distortion} shows an
importance sampling estimate ($1024$ samples per test example were
taken) of the negative log-likelihood for the several methods,
together with the converged values of each method in the RD plane.
Both show results for which the bit-vector has dimensionality $D=32$
and $D=128$. Regarding the estimated negative log-likelihood, our
methods exhibit increased performance when compared to SFE, and
top-$k$ sparsemax is competitive with the remaining unbiased
estimators. However, in the RD plane, both our methods show
comparable performance to SFE$+$ and NVIL for $D=32$, but for $D=128$
all of our methods have a significantly higher rate and lower
distortion than any unbiased estimator, suggesting a better fit of
$p(x|\phi)$~\citep{Alemi2018}. In \figref{fig:structcalls}, we can observe the training
progress in number of calls to $p(x \mid z, \phi)$ for the models
with 32 and 128 latent bits, respectively. While $\sparsemaxk$
introduces bias towards the most probable assignments and may discard
outcomes that $\sparsemax$ would assign non-zero probability to, as
training progresses distributions may (or tend to) be sufficiently
sparse and this mismatch disappears, making the gradient computation
exact. Remarkably, this happens for $D=32$\,---\,the support of
$\sparsemaxk$ is smaller than $k$, giving the true gradient to
$q(z\mid x, \lambda)$ for most of the training. This no longer
happens for $D=128$, for which it remains with full support
throughout, due to the much larger search space. On the other hand, \smap
solutions become very sparse from the start in both cases, while
still obtaining good performance. There is, therefore, a trade-off
between the solutions we propose: on one hand, $\sparsemaxk$ can
become exact with respect to the expectation in \eqnref{eq:fit}, but
it only does so if the chosen $k$ is suitable to the difficulty of
the model; on the other hand, \smap may not offer an exact gradient
to $q(z\mid x, \lambda)$, but its performance is very close to
$\sparsemaxk$ and its higher propensity for sparsity gifts it with
less computation (App.~\ref{app:dec_calls}).

Concerning relaxed estimators, note that the reconstruction loss is
computed given a continuous sample, rather than a discrete one,
allowing it more flexibility to directly reduce distortion and
potentially explaining why it does well in that regard. Moreover, the
rate of the relaxed model is unknown,\footnote{Estimating it would
require a choice of Binary Concrete prior and an estimate of the KL
divergence from that to the Binary Concrete approximate posterior
\citep[Appendix C.3.2]{Concrete}.} and instead we plot the rate as if
$z$ was given discrete treatment, which, although common practice,
makes comparisons to other estimators inadequate. For ST
Gumbel-Softmax the situation is different since, after training, $z$
is given discrete treatment throughout. Its success shows that,
unlike in the other tasks considered, training on biased gradients is
not too problematic.

\section{Related Work}

\paragraph{Differentiable sparse mappings.} There has been recent interest in
applying sparse mappings of discrete distributions in deep discriminative
models~\citep{martins2016softmax, niculae2018sparsemap, fusedmax,
entmax, sparsemapcg}, attention mechanisms~\citep{malaviya2018sparse, shao2019ssn,
maruf2019selective, correia2019adaptively}, and in
topic models~\citep{caothesis}. Our work focuses
on the parameterization of distributions over latent variables with
sparse mappings, on the computational advantage to be gained by sparsity, 
and on the contrast between our novel training
method and common sampling-based methods.

\paragraph{Reducing sampling noise.} The sampling procedure found in
SFE is a great source of variance in models that build upon this
method. To reduce this variance, many works have proposed
baselines~\citep{Williams1992,MuProp,CV2013}. VIMCO~\citep{VIMCO} is
a multi-sample estimator which exploits variance reduction via
input-dependent baselines as well as a lower bound on marginal
likelihood which is tighter than the ELBO~\citep{IWAE}. The number of
samples in VIMCO is a hyperparameter that stays fixed throughout
training. Our methods, in contrast, may take several decoder calls
initially, but that number automatically decreases over time as
training progresses. While baselines
must be independent of the sample for which we assess the score
function, exploiting correlation in the downstream losses of
dependent samples holds potential for further variance reduction.
These are known as control variates~\citep{GreensmithEtAl}.
REBAR~\citep{REBAR} exploits a continuous relaxation to obtain a
dependent sample and uses the downstream loss assessed at the relaxed
sample to define a control variate. RELAX~\citep{RELAX}, instead,
learns to predict the downstream loss of the relaxed sample with
an auxiliary network.
In contrast, sparse marginalization
works for any factorization where a primitive for $1$-best (or
$k$-best) enumeration is available, and takes no additional
parameters nor additional optimization objectives.
Another line of work approximates argmax gradients by
perturbed finite differences
\cite{lorberbom2019direct,vlastelica};
this requires the same computation primitive as our approach,
but is always biased.
ARM~\citep{yin2019arm} is a control variate
based on antithetic samples~\citep{mcbook}: it does not require relaxation nor
additional parameters, but it only applies to
factorial Bernoulli distributions.
Closest to our work are variance reduction techniques that rely on partial
marginalization, typically of the top-$k$ assignments to the latent
variable~\citep{RB19,Kool2020Estimating}. These methods show improved
performance and variance reduction, but
require rejection sampling, which can be challenging in structured problems.

\section{Conclusion}
We described a novel training strategy for discrete latent variable
models, eschewing the common approach based on MC gradient estimation
in favor of deterministic, exact marginalization under a sparse
distribution. Sparsity leads to a powerful \emph{adaptive} method,
which can investigate fewer or more latent assignments $z$ depending
on the ambiguity of a training instance $x$, as well as on the stage
in training.
We showcase the performance and flexibility of our method by
investigating a variety of applications, with both discrete and
structured latent variables, with positive results. Our models very
quickly approach a small number of latent assignment evaluations per
sample, but make progress much faster and overall lead to superior
results.
Our proposed method thus offer the accuracy and robustness of exact
marginalization while meeting the efficiency and flexibility of score
function estimator methods, providing a promising alternative.

\section*{Broader Impact}

We discuss here the broader impact of our work. Discussion in this
section is predominantly speculative, as the methods described in
this work are not yet tested in broader applications. However, we do
think that the methods described here can be applied to many
applications\,---\,as this work is applicable to any model that
contains discrete latent variables, even of combinatorial type.

Currently, the solutions available to train discrete latent variable
models (LVMs) greatly rely on MC sampling, which can have high variance.
Methods that aim to decrease this variance are often not trivial to
train and to implement and may disincentivize practitioners from
using this class of models. However, we believe that discrete LVMs
have, in many cases, more interpretable and intuitive
latent representations. Our methods offer: a simple approach in
implementation to train these models; no addition in the number of
parameters; low increase in computational overhead (especially when
compared to more sophisticated methods of variance
reduction~\citep{RB19}); and improved performance. Our code has been
open-sourced as to ensure it's scrutinizable by
anyone and to boost any related future work that other researchers
might want to pursue.

As we have already pointed out, oftentimes LVMs
have superior explanatory power and so can aid in understanding cases
in which the model failed the downstream task. Interpretability of
deep neural models can be essential to better discover any ethically
harmful biases that exist in the data or in the model itself.

On the other hand, the generative models discussed in this work may
also pave the way for malicious use cases, such as is the case with
\emph{Deepfakes}, fake human avatars used by malevolent Twitter
users, and automatically generated fraudulent news. Generative models
are remarkable at sampling new instances of fake data and, with the
power of latent variables, the interpretability discussed before can
be used maliciously to further push harmful biases instead of
removing them. Furthermore, our work is promising in improving the
performance of LVMs with several discrete
variables, that can be trained as attributes to control the sample
generation. Attributes that can be activated or deactivated at will
to generate fake data can both help beneficial and malignant users to
finely control the generated sample. Our work may be currently
agnostic to this, but we recognize the dangers and dedicate effort to
combating any malicious applications.

Energy-wise, LVMs often require less data and
computation than other models that rely on a massive amount of data
and infrastructure. This makes LVMs ideal for
situations where data is scarce, or where there are few computational
resources to train large models. We believe that better latent
variable modeling is a step forward in the direction of alleviating
environmental concerns of deep learning
research~\citep{strubell2019energy}. However, the models proposed in
this work tend to use more resources earlier on in training than
standard methods, and even though in the applications shown they
consume much less as training progresses, it's not clear if that
trend is still observed in all potential applications.

In data science, LVMs, such as
mixed-membership models~\citep{blei2014build}, can be used to uncover
correlations in large amounts of data, for example, by clustering
observations. Training these models requires various degrees of
approximations which are not without consequence, they may impact the
quality of our conclusions and their fealty to the data. For example,
variational inference tends to under-estimate uncertainty and give
very little support to certain less-represented groups of variables.
Where such a model informs decision-makers on matters that affect
lives, these decisions may be based on an incomplete view of the
correlations in the data and/or these correlations may be exaggerated
in harmful ways. On the one hand, our work contributes to more stable
training of LVMs, and thus it is a step towards addressing some of
the many approximations that can blur the picture. On the other hand,
sparsity may exhibit a larger risk of disregarding certain
correlations or groups of observations, and thus contribute to
misinforming the data scientist. At this point it is unclear to which
extent the latter happens and, if it does, whether it is consistent
across LVMs and their various uses. We aim to study this issue
further and work with practitioners to identify failure cases.

\begin{ack}

Top-$k$ sparsemax is due in great part to initial work and ideas of
Mathieu Blondel. The authors are thankful to Wouter Kool for feedback
and suggestions. We are grateful to Ben Peters, Erick Fonseca, Marcos
Treviso, Pedro Martins, and Tsvetomila Mihaylova for insightful group
discussion. We would also like to thank the anonymous reviewers for
their helpful feedback.

This work was partly funded by the European Research Council (ERC StG
DeepSPIN 758969), by the P2020 project MAIA (contract 045909), and by
the Fundação para a Ciência e Tecnologia through contract
UIDB/50008/2020. This work also received funding from the European
Union’s Horizon 2020 research and innovation programme under grant
agreement 825299 (GoURMET).

\end{ack}

\bibliography{paper}
\bibliographystyle{unsrtnat}

\newpage
\onecolumn
\appendix

\section{Computing sparsemax: Forward and Backward Passes}\label{app:sparsemax}
The sparsemax mapping \cite{martins2016softmax}, as discussed in
Section~\ref{sec:categorical}, is given by the unique solution to
\begin{equation}\label{eq:sparsemax_supp}
    \sparsemax(\s) \coloneqq \argmin_{\pv \in \simplex^K} \| \pv - \s \|_2^2\,.
\end{equation}
As a projection onto a polytope, the solution is likely to fall on the
boundaries or the corners of the set. In this case, points on the boundary of
$\simplex^K$ have one or more zero coordinates. In contrast, $\softmax(\s)
\propto \exp(\s)$ is always strictly inside the simplex.
From the optimality conditions of the sparsemax problem~\eqref{eq:sparsemax_supp},
it follows that the solution must have the form:
\begin{equation}\label{eq:sparsemax_form}
\sparsemax(\s) = \max(\s - \tau, 0)\,,
\end{equation}
where the maximum is elementwise, and $\tau$ is the unique value that
ensures the result sums to one.
Letting $\bar{\ZZ}$ be the set of nonzero coordinates in the solution,
the normalization condition is equivalently
\begin{equation}
\tau = \frac{\sum_{z \in \bar{\ZZ}} \ss_z}{|\bar{\ZZ}|}\,.
\end{equation}
Observing that small changes to $\s$ almost always have no effect on the support
$\bar{\ZZ}$, differentiating Equation~\ref{eq:sparsemax_form} gives
\begin{equation}
\pfrac{\bar{\pv}}{\bar{\s}} = \bm{I}_{|\bar{\ZZ}|} - \frac{1}{|\bar{\ZZ}|}
\bm{11}^\top\,,
\end{equation}
where $\bar{\pv}$ and ${\bar{\s}}$ denote the subsets of the respective vectors
indexed by the support $\bar{\ZZ}$. Outside of the support, the partial
derivatives are zero. (\emph{Cf.} the more general result in \cite[Proposition
2]{entmax}.)
In terms of computation, $\tau$ may be found numerically using root finding
algorithms on $f(\tau) = \max(\s - \tau, 0) - 1$.
Alternatively, observe that it is enough to find $\bar{\ZZ}$. By showing that
sparsemax must preserve the ordering, \ie, that if $\ss_{z'} > \ss_z$ and $z \in
\bar{\ZZ}$ then $z' \in \bar{\ZZ}$, it can be shown that $\bar{\ZZ}$ must
consist of the $k$ highest-scoring coordinates of $\s$, where $k$ can be find by
inspection after sorting $\s$. This leads to a straightforward $\mathcal{O}(K
\log K)$ algorithm due to Held \citep[pp.~16--17]{Held1974}. This can be further pushed to
$\mathcal{O}(K)$ using median pivoting algorithms~\cite{Condat2016}. We use
a simpler implementation based on repeatedly calling $\topk$,
doubling $k$ until the optimal solution is found. Since solutions get sparser
over time and $\topk$ is GPU-accelerated
in modern libraries \cite{pytorch}, this strategy is very fast in practice.

\section{Computing SparseMAP: The Active Set Algorithm}\label{app:activeset}
In this section, we present the active set method
\citep[Chapters 16.4 \& 16.5]{nocedalwright} as applied to the SparseMAP
optimization problem (Eq.~\ref{eq:sparsemap})~\cite{niculae2018sparsemap}.
This form of the algorithm, due to Martins et al.~\cite[Section 6]{ad3},
is a small variation of the formulation of Nocedal and Wright for handling
the equality constraint. Recall the SparseMAP problem,
\begin{equation}\label{eq:sparsemap_supp}
\argmin_{\pv \in \simplex^{|\ZZ|}} \|\bm{A}\pv - \bm{t}\|_2^2\,.
\end{equation}
Assume that we could identify the \emph{support}, or
\emph{active set} of an optimal solution  $\bm{\xi}^\star$, denoted
\[\bar\ZZ \defeq \left\{z \in \ZZ \mid \xi^\star_z > 0\right\}\,. \]
Then, given this set, we could find the solution to \eqref{eq:sparsemap_supp}
by solving the lower-dimensional equality-constrained problem
\begin{equation}
\minimize\| \bar{\bm{A}}\bar{\pv} - \bm{t} \|^2
\quad\subjto\quad\bm{1}^\top\bar{\pv} = 1\,,
\end{equation}
where we denote by $\bar{\bm{A}}$ and $\bar{\pv}$ the restrictions of $\bm{A}$
and $\pv$ to the active set of structures $\bar\ZZ$.
The solution to this equality-constrained QP satisfies the KKT optimality
conditions,
\begin{equation}
    \label{eq:qp_kkt}
    \begin{bmatrix}
        {\bar{\bm{A}}}^\top \bar{\bm{A}} & \bm{1} \\
        \bm{1}^\top & 0 \\
    \end{bmatrix}
    \begin{bmatrix} \bar{\pv} \\ \tau \end{bmatrix}
    =
    \begin{bmatrix} \bar{\bm{A}}^\top \bm{t} \\ 1 \end{bmatrix}.
\end{equation}
Of course, the optimal support is not known ahead of time. The active set
algorithm attempts to guess the support in a greedy fashion,
at each iteration either [if the solution of \eqref{eq:qp_kkt} is not feasible
for \eqref{eq:sparsemap_supp}]
dropping a structure from $\bar\ZZ$, or [otherwise] adding a new structure.
Since the support changes one structure at a time, the design matrix in
\eqref{eq:qp_kkt} gains or loses one row and column, so
we may efficiently maintain its Cholesky decomposition via rank-one updates.

We now give more details about the computation.
Denote the solution of Eq.~\ref{eq:qp_kkt}, (extended with zeroes),
by $\hat{\pv} \in \simplex^{|\ZZ|}$.
Since we might not have the optimal $\bar{\ZZ}$ yet, $\hat{\pv}$ can be infeasible
(some coordinates may be negative.)
To account for this, we take a partial step in its direction,
\begin{equation}
\pv^{(i+1)} = (1-\gamma)\pv^{(i)} + \gamma \hat{\pv}^{(i+1)}\,
\end{equation}
where, to ensure feasibility, the step size is given by
\begin{equation}\label{eq:step_size}
\gamma = \min \left(1, \min_{z \in \bar{\ZZ}; \pp^{(i)}_z > \hat{\pp}_z}
\frac{
\pp^{(i)}_z
}{
\pp^{(i)}_z - \hat{\pp}_z}
\right)\,.
\end{equation}

If, on the other hand, $\hat{\pv}$ is feasible for \eqref{eq:sparsemap_supp},
(so $\gamma=1$),
we check whether we have a globally optimal solution.
By construction, $\hat{\pv}$ satisfies all KKT
conditions except perhaps dual feasibility $\bm{\nu} \geq 0$,
where $\nu_z$ is the dual variable (Lagrange multiplier) corresponding to the
constraint $\pp_z \geq 0$.
Denote $\bm{\mu}^{(i)} \defeq \bm{A}\pv^{(i)} = \bar{\bm{A}}\bar{\pv}^{(i)}$.
For any $z \not\in \bar{\ZZ}$, the corresponding dual variable must satisfy
\begin{equation}\label{eq:dual}
\nu_z = \tau^{(i)} - \DP{\bm{a}_z}{\bm{t} - \bm{\mu}^{(i)}}\,.
\end{equation}
If the smallest dual variable is positive, then our current guess satisfies all
optimality conditions. To find the smallest dual variable we can equivalently
solve $\argmax_{z \in \ZZ} \DP{\bm{a}_z}{\bm{t} - \bm{\mu}^{(i)}}$, which is
a maximization (MAP) oracle call. If the resulting $\nu_z$ is negative,
then $z$ is the index of the most violated constraint $\pp_z \geq 0$;
it is thus a good choice of structure to add to the active set.

The full procedure is given in Algorithm~\ref{alg:activeset}.
The backward pass can be computed by implicit differentiation of the KKT system \eqref{eq:qp_kkt} \wrt
$\bm{t}$, giving, as in \cite{niculae2018sparsemap},
\begin{equation}
\pfrac{\bar{\pv}}{\bm{t}} = \bar{\bm{A}}\big(\bm{S} - \bm{ss}^\top / s\big),
\quad\text{where}\quad
\bm{S} = (\bar{\bm{A}}^\top \bar{\bm{A}})^{-1},
\bm{s} = \bm{S1}, s = \bm{1}^\top\bm{S1}
\,.
\end{equation}
It is possible to apply the $\ell_2$ regularization term only to a subset of the
rows of $\bm{A}$, as is more standard in the graphical model literature. We
refer the reader to the presentation in \cite{ad3,niculae2018sparsemap} for this extension.

\begin{algorithm}[h!]\small
\caption{Active set algorithm for SparseMAP \label{alg:activeset}}
\begin{algorithmic}[1]
\Statex \textbf{Init:}
$\bar\ZZ^{(0)} = \{ z^{(0)}\}$
\quad where \quad
$z^{(0)} \in \argmax_{z \in \ZZ} \DP{\bm{a}_z}{\bm{t}}$
or a random structure.
\For {i \text{in} $1, \ldots, N$}
\State Compute $\tau^{(i)}$ and $\hat\pv^{(i)}$ by solving the relaxed QP (Eq.~\ref{eq:qp_kkt}).
\Comment{Cholesky update.}
\State $\pv^{(i)} \gets (1-\gamma) \pv^{(i-1)} + \gamma \hat{\pv}^{(i)}$
~(with $\gamma$ from Eq.~\ref{eq:step_size}).
\If{$\gamma < 1$}
\State Drop the minimizer of Eq.~\ref{eq:step_size} from $\bar{\ZZ}^{(i)}$.
\Else
\State Find most violated constraint,
$z^{(i)} \gets \argmin_{z \in \ZZ} \nu_z$.
\Comment{Eq.~\ref{eq:dual}, MAP oracle.}
\If{$\nu_{z^{(i)}} \geq 0$}
\State \Return \Comment{Converged.}
\Else
\State $\mathcal{Z}^{(i+1)} \gets \mathcal{Z}^{(i)} \cup \{ z^{(i)} \}$
\EndIf
\EndIf
\EndFor
\end{algorithmic}
\end{algorithm}

\section{Budget Constraint}\label{app:budget}

The maximization oracle for the budget constraint described
in~\secref{bernvae} can be computed in $\mathcal{O}(D \log D)$. This
is done by sorting the Bernoulli scores and selecting the entries
among the top-$B$ which have a positive score.

\section{Importance Sampling of the Marginal Log-Likelihood}\label{app:is}

Bits-per-dimension is the negative logarithm of marginal
likelihood normalized per number of pixels in the image, thus we need
to assess or estimate the marginal likelihood of observations. For
dense parameterizations, the usual option is importance sampling (IS)
using the trained approximate posterior as importance distribution:
\ie, $\log p(x|\phi) \overset{\text{IS}}{\approx} \log
\left(\frac{1}{S} \sum_{s=1}^S \frac{p(z^{(s)}, x|\phi)}{q(z^{(s)} |
x, \lambda)} \right)$ with $z^{(s)} \sim q(z|x, \lambda)$. The result
is a stochastic lower bound which converges to the true log-marginal
in the limit as $S \to \infty$. With a sparse posterior approximation
we can split the marginalization
\begin{equation}
\log p(x|\phi) = \log \left( \sum_{z \in \bar \ZZ} p(z)p(x|z, \phi) + \sum_{z \in \ZZ \setminus \ZZ} p(z)p(x|z, \phi)  \right)
\end{equation}
into one part that handles outcomes in the support $\bar \ZZ$ of the
sparse posterior approximation and another part that handles the
outcomes in the complement set $\ZZ \setminus \bar \ZZ$. We compute
the first part exactly and estimate the second part via rejection
sampling from $p(z)$.

\section{Training Details}\label{app:details}

In our applications, we follow the experimental procedures described
in~\citep{RB19} and~\citep{Lazaridou2017} for~\secref{gen} and
\secref{comm}, respectively. We describe below the most relevant
training details and key differences in architectures when
applicable. For other implementation details that we do not mention
here, we refer the reader to the works referenced above. For all
Gumbel baselines, we relax the sample into the continuous space but
assume a discrete distribution when computing the entropy of $\pi(z
\mid x, \theta)$, as suggested as one implementation option in
\citet{Concrete}. Our code is publicly available at
\href{https://github.com/deep-spin/sparse-marginalization-lvm}{\tt
https://github.com/deep-spin/sparse-marginalization-lvm} and was
largely inspired by the structure and implementations found in
EGG~\citep{Kharitonov2019} and was built upon it.

\paragraph{Semisupervised Variational Autoencoder.} In this
experiment, the classification network consists of three fully
connected hidden layers of size 256, using ReLU activations. The
generative and inference network both consist of one hidden layer of
size 128, also with ReLU activations. The multivariate Gaussian has 8
dimensions and its covariance is diagonal. For all models we have
chosen the learning rate based on the best ELBO on the validation
set, doing a grid search (5e-5, 1e-4, 5e-4, 1e-3, 5e-3). The accuracy
shown in \figref{fig:ssvaeelbo} is the test accuracy taken after the last
epoch of training. The temperature of the Gumbel models was annealed
according to $\tau = \max\left(0.5, -rt\right)$, where $t$ is the
global training step. For these models, we also did a grid search
over $r$ (1e-5, 1e-4) and over the frequency of updating $\tau$ every
(500, 1000) steps. Optimization was done with Adam. For our method,
in the labeled loss component of the semisupervised objective we used
the sparsemax loss~\citep{martins2016softmax}. Following
\citet{RB19}, we pretrain the network with only labeled data prior to
training with the whole training set. Likewise, for our method, we
pretrained the network on the sparsemax loss and every other method
with the Negative Log-Likelihood loss.

\paragraph{Emergent communication game.} In this application, we
closely followed the experimental procedure described
by~\citet{Lazaridou2017} with a few key differences. The architecture
of the sender and the receiver is identical with the exception that
the sender does not take as input the distracting images along with
the correct image\,---\,only the correct image. The collection of
images shown to the receiver was increased from 2 to 16 and the
vocabulary of the sender was increased to 256. The hidden size and
embedding size was also increased to 512 and 256, respectively. We
did a grid search on the learning rate (0.01, 0.005, 0.001) and
entropy regularizer (0.1, 0.05, 0.01) and chose the best
configuration for each model on the validation set based on the
communication success. For the Gumbel models, we applied the same
schedule and grid search to the temperature as described for
Semisupervised VAE. All models were trained with the Adam optimizer,
with a batch size of 64 and during 200 epochs. We choose the
vocabulary of the sender to be 256, the hidden size to be 512 and the
embedding size to be 256.

\paragraph{Bit-Vector Variational Autoencoder.} In this experiment,
we have set the generative and inference network to consist of one
hidden layer with 128 nodes, using ReLU activations. We have searched
a learning rate doing grid search (0.0005, 0.001, 0.002) and chosen
the model based on the ELBO performance on the validation set. For
the Gumbel models, we applied the same schedule and grid search to
the temperature as described for Semisupervised VAE. We used the Adam
optimizer.

\subsection{Datasets}

\paragraph{Semisupervised Variational Autoencoder.} MNIST consists of
$28 \times 28$ gray-scale images of hand-written digits. It contains
60,000 datapoints for training and 10,000 datapoints for testing. We
perform model selection on the last 10,000 datapoints of the training
split.

\paragraph{Emergent communication game.} The data used by
\citet{Lazaridou2017} is a subset of ImageNet containing 463,000
images, chosen by sampling 100 images from 463 base-level concepts.
The images are then applied a forward-pass through the pretrained
VGG ConvNet~\citep{convnet} and the representations at the
second-to-last fully connected layer are saved to use as input to the
sender/receiver.

\paragraph{Bit-Vector Variational Autoencoder.} Fashion-MNIST
consists of $28 \times 28$ gray-scale images of clothes. It contains
60,000 datapoints for training and 10,000 datapoints for testing. We
perform model selection on the last 10,000 datapoints of the training
split.

\section{Performance in Decoder Calls}\label{app:dec_calls}

\begin{figure*}[ht]
    \centering
    \includegraphics[width=0.45\columnwidth]{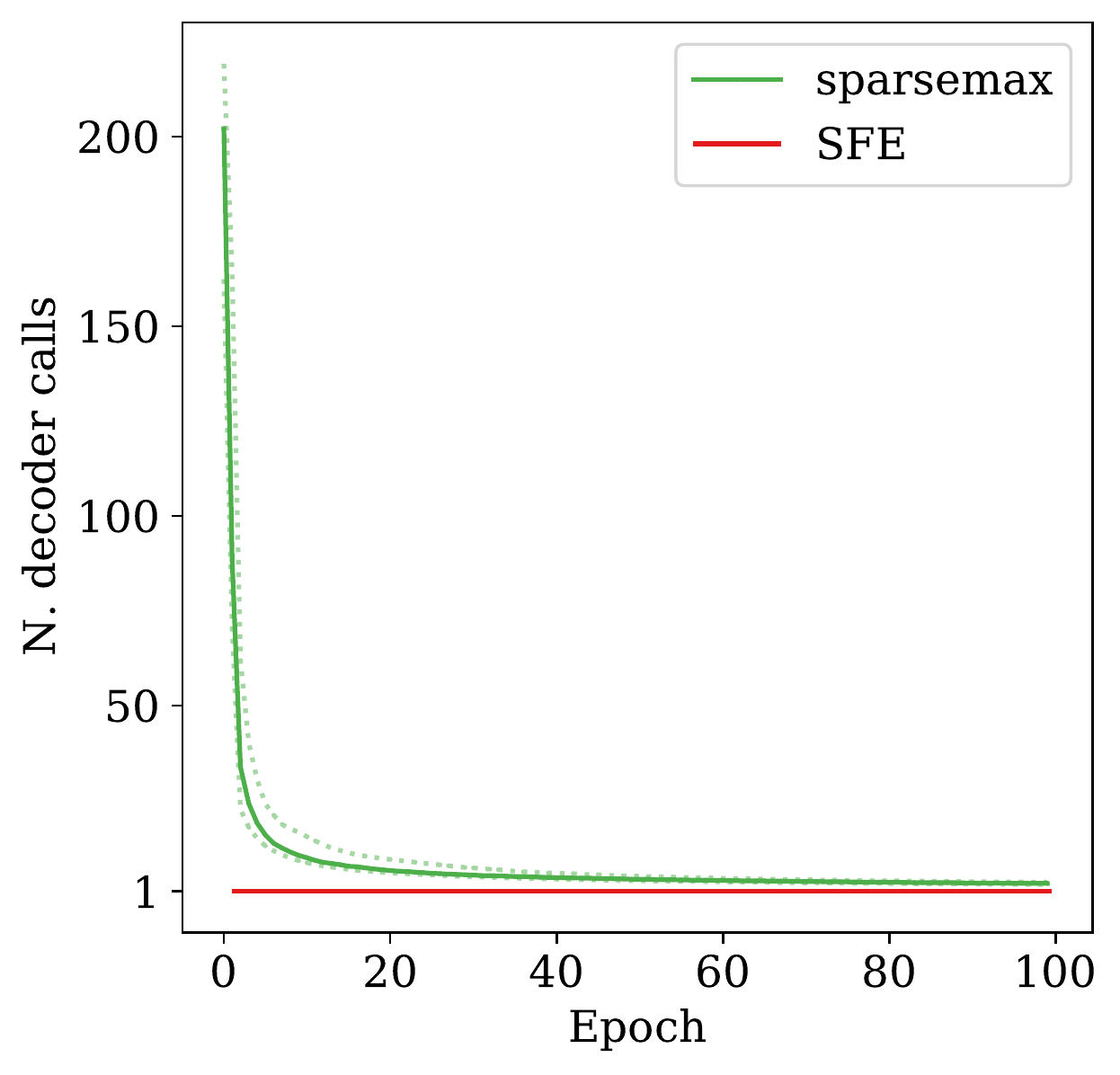}
    \caption{
        Median decoder calls per epoch during training
        time with 10 and 90 percentiles in dotted lines by sparsemax in \secref{comm}.
    }
    \label{fig:nonzero_comm}
\end{figure*}

\figref{fig:nonzero_comm} shows the number of decoder calls with
percentiles for the experiment in \secref{comm}. While dense right at the beginning
of training, support quickly falls to an average of close to $1$ decoder call.

\begin{figure}[ht]
    \centering
    \begin{subfigure}[b]{0.45\textwidth}
        \centering
        \includegraphics[width=\textwidth]{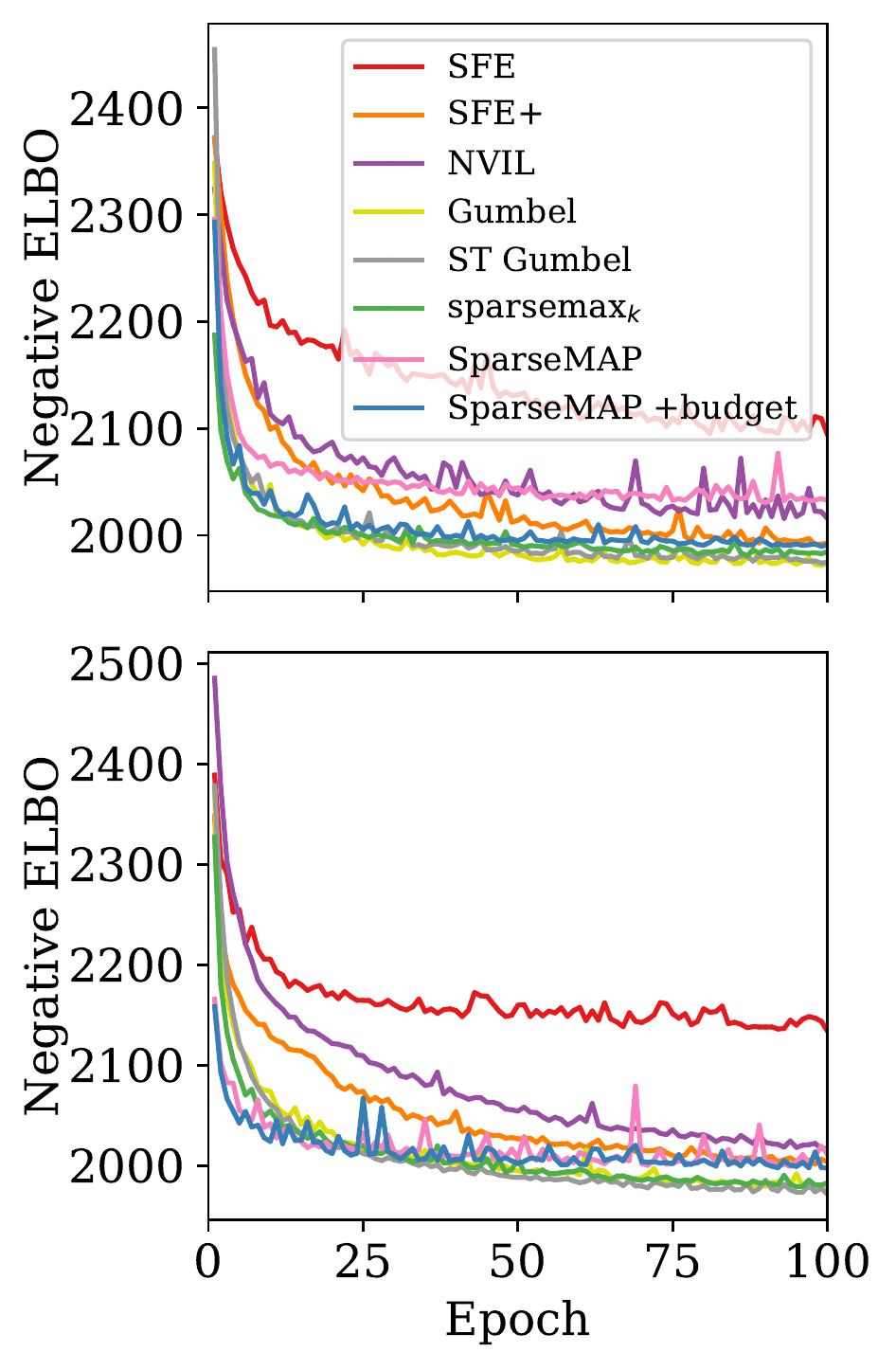}
        \caption{Negative ELBO over training epochs.}
        \label{fig:elbo_bit_epochs}
    \end{subfigure}
    \hfill
    \begin{subfigure}[b]{0.45\textwidth}
        \centering
        \includegraphics[width=\textwidth]{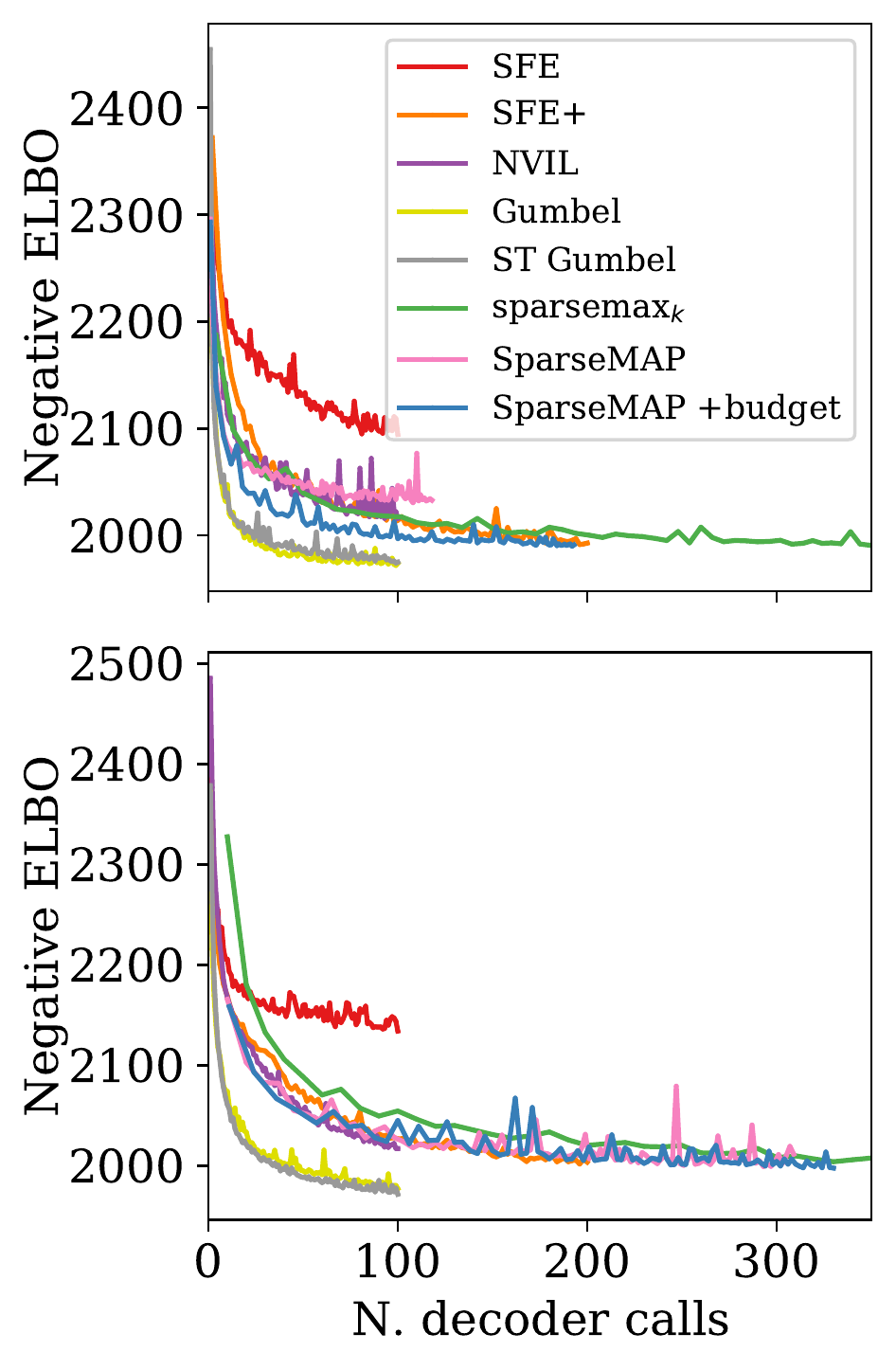}
        \caption{Negative ELBO over decoder calls.}
        \label{fig:elbo_bit_calls}
    \end{subfigure}
    \caption{Performance on the validation set
    for the experiment in \secref{bernvae},
    $D=32$ (top) / $D=128$ (bottom). For $D=32$, top-$k$ sparsemax continues
    until a total of 561 median decoder calls, and for $D=128$ it continues
    until a total of 998.}
    \label{fig:elbo_bit}
\end{figure}

\figref{fig:elbo_bit}
shows the downstream loss (ELBO) of experiment \secref{bernvae} over epochs
and over the median number of decoder calls per epoch. The plots on \figref{fig:elbo_bit_calls}
show how our methods have comparable computational overhead to sampling approaches. Oftentimes,
our methods could have been trained in less epochs to obtain the same performance as the
sampling estimators have for 100 epochs.

\section{Computing infrastructure}

Our infrastructure consists of 4 machines with the specifications
shown in Table~\ref{table:computing_infrastructure}. The machines
were used interchangeably, and all experiments were executed in a
single GPU. Despite having machines with different specifications, we
did not observe large differences in the execution time of our models
across different machines.

\begin{table}[!hb]
    \small
    \begin{center}
    \begin{tabular}{l ll}
        \toprule
        \sc \# & \sc GPU & \sc CPU  \\
        \midrule
        1.   & 4 $\times$ Titan Xp - 12GB           & 16 $\times$ AMD Ryzen 1950X @ 3.40GHz - 128GB \\
        2.   & 4 $\times$ GTX 1080 Ti - 12GB        & 8 $\times$ Intel i7-9800X @ 3.80GHz - 128GB \\
        3.   & 3 $\times$ RTX 2080 Ti - 12GB        & 12 $\times$ AMD Ryzen 2920X @ 3.50GHz - 128GB \\
        4.   & 3 $\times$ RTX 2080 Ti - 12GB        & 12 $\times$ AMD Ryzen 2920X @ 3.50GHz - 128GB \\
        \bottomrule
    \end{tabular}
    \end{center}
    \caption{Computing infrastructure.}
    \label{table:computing_infrastructure}
\end{table}

\end{document}